\documentclass[conference]{IEEEtran}
\IEEEoverridecommandlockouts
\usepackage{cite}
\usepackage{amsmath,amssymb,amsfonts}
\usepackage{algorithmic}
\usepackage{graphicx}
\usepackage{textcomp}
\usepackage{xcolor}
\def\BibTeX{{\rm B\kern-.05em{\sc i\kern-.025em b}\kern-.08em
    T\kern-.1667em\lower.7ex\hbox{E}\kern-.125emX}}
\newcommand{\etal}{\textit{et al.}}
\newcommand{\ie}{\textit{i.e.}}
\newcommand{\eg}{\textit{e.g.}}
\usepackage{booktabs}
\usepackage{caption}
\usepackage{hyperref}

\begin{document}

\title{EmoStory: Emotion-Aware Story Generation}

\author{
    Jingyuan Yang, Rucong Chen, Weibin Luo, Hui Huang\textsuperscript{*} \\
    CSSE, Shenzhen University
}


\twocolumn[{
	\renewcommand\twocolumn[1][]{#1}
	\maketitle
	\begin{center}
		\centering
		\includegraphics[width=\linewidth]{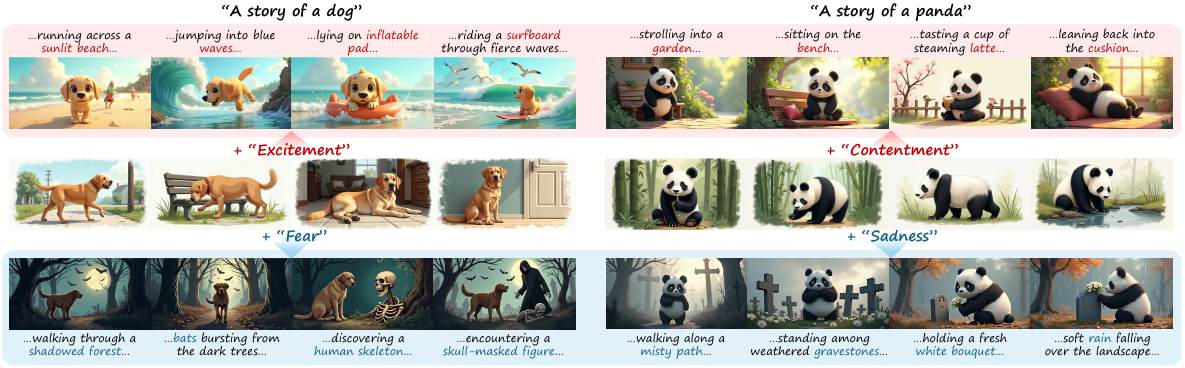}
		\captionof{figure}{Emotion-aware story generation with EmoStory, which introduces emotions (top: \textit{positive}, bottom: \textit{negative}) to given subjects (middle: \textit{neutral}), generating coherent and emotionally expressive visual stories.}
		\label{fig:teaser}
	\end{center}
}]

\begin{abstract}
Story generation aims to produce image sequences that depict coherent narratives while maintaining subject consistency across frames.
Although existing methods have excelled in producing coherent and expressive stories, they remain largely emotion-neutral, focusing on \textit{what} subject appears in a story while overlooking \textit{how} emotions shape narrative interpretation and visual presentation.
As stories are intended to engage audiences emotionally, we introduce emotion-aware story generation, a new task that aims to generate subject-consistent visual stories with explicit emotional directions.
This task is challenging due to the abstract nature of emotions, which must be grounded in concrete visual elements and consistently expressed across a narrative through visual composition. 
To address these challenges, we propose EmoStory, a two-stage framework that integrates agent-based story planning and region-aware story generation. 
The planning stage transforms target emotions into coherent story prompts with emotion agent and writer agent, while the generation stage preserves subject consistency and injects emotion-related elements through region-aware composition.
We evaluate EmoStory on a newly constructed dataset covering 25 subjects and 600 emotional stories. 
Extensive quantitative and qualitative results, along with user studies, show that EmoStory outperforms state-of-the-art story generation methods in emotion accuracy, prompt alignment, and subject consistency.
\end{abstract}


\section{Introduction}
\label{sec:intro}

Story generation aims to generate an image sequence that depicts a narrative while maintaining subject consistency and inter-frame coherence.
With the rapid development of diffusion-based generative models~\cite{rombach2022high}, recent methods have achieved substantial progress in producing more continuous, detailed, and expressive story images~\cite{zhou2024storydiffusion,liu2024intelligent,tewel2024training}.
However, these approaches remain largely emotion-neutral: they model \textit{what} subjects occur in the narrative, yet overlook \textit{how} they should be emotionally interpreted or visually enhanced.

Stories are created to affect and engage their readers, making emotional control essential in story generation~\cite{parkinson1993making}.
Psychological and artistic studies show that emotion shapes narrative themes and events, influences the selection of elements, and guides visual cues such as color, lighting, and composition~\cite{yang2023emoset}.
Two stories that share identical subjects may evoke entirely different emotions.
For example, in Fig.~\ref{fig:teaser}, given the subject ``dog'', adding elements such as waves and a surfboard can create \textit{excitement}, whereas introducing bats or a human skeleton can shift the story toward \textit{fear}.
This raises a key question: can we automatically generate visual stories with specific emotional directions?

In this paper, we introduce a new task, emotion-aware story generation, which aims to generate image sequences that are not only subject-consistent but also emotion-evoking, as illustrated in Fig.~\ref{fig:teaser}.
However, incorporating emotions into story generation presents several challenges.
First, emotions are abstract concepts that must be grounded in concrete visual elements (sunlit beach, waves) to form a coherent narrative.
Second, this grounding is difficult since emotional impact relies heavily on visual presentations such as semantic clarity and composition, which text alone cannot fully specify.

To address these challenges, we propose EmoStory, a framework that introduces emotional expressiveness into story generation through two complementary stages: agent-based story planning and region-aware story generation.
In the first stage, two collaborative agents, \ie, an emotion agent and a writer agent, work together to construct the narrative foundation.
We first build emotion factor trees from EmoSet to map each target emotion to representative visual elements.
Using this knowledge, emotion agent produces a story script that specifies the subject, emotional elements, theme, and event, following a standard narrative structure.
Writer agent further refines and connects these components into a series of coherent story prompts with emotional expressiveness.
This planning stage transforms abstract emotions into concrete narrative prompts.
In the second stage, to express emotion while maintaining subject consistency, we introduce a region-aware story generation module that separates each image into subject and non-subject regions.
Specifically, we design a subject alignment module to preserve subject consistency within subject region and an emotion composition module to inject emotional elements into non-subject regions via cross-attention.
This design ensure accurate subject depiction and emotional composition.
Through this two-stage design, EmoStory generates visual stories that are not only subject-consistent but also emotionally evocative.

EmoStory is evaluated on a constructed dataset containing 25 subjects and 8 emotion directions, generating 3 stories for each subject–emotion pair, for a total of 600 stories.
We compare our method with state-of-the-art story generation approaches using three metrics: Emotion Accuracy, Prompt Alignment, and Subject Consistency.
Both quantitative and qualitative results demonstrate the superiority of EmoStory, while user studies and visualizations further confirm its human alignment and effectiveness.

In summary, our contributions are:
\begin{itemize}
	\setlength{\itemsep}{0pt}
	\setlength{\parsep}{0pt}
	\setlength{\parskip}{0pt}
	
	\item We introduce emotion-aware story generation, a new task that aims to generate subject-consistent image sequences with explicit emotional directions, enabling visual stories narratively coherent and emotionally evocative.
	
	\item We propose EmoStory, a two-stage framework that integrates agent-based story planning and region-aware story generation, effectively grounding emotions into concrete narrative prompts and expressive visual compositions.
	
	\item We conduct extensive experiments, demonstrating that EmoStory outperforms state-of-the-art methods in emotion accuracy, prompt alignment, and subject consistency.

\end{itemize}

\section{Related work}
\label{sec:rw}

\subsection{Story Generation}

With the rapid advancement of text-to-image (T2I) generation, spanning diffusion models~\cite{rombach2022high}, DiTs~\cite{peebles2023scalable}, and autoregressive generators~\cite{sun2024autoregressive}, story generation has progressed in parallel~\cite{zhou2024storydiffusion,dinkevich2025story2board,liu2024intelligent,ye2023ip,cai2025diffusion,tewel2024training,liu2025one,he2025dreamstory}.
StoryDiffusion~\cite{zhou2024storydiffusion} improves image consistency via consistent self-attention, while Story2Board~\cite{dinkevich2025story2board} introduces latent panel anchoring and reciprocal attention value mixing.
StoryGen~\cite{liu2024intelligent} further explores learning-based autoregressive image generation with a visual language context module.
Unlike text-to-image generation, story generation additionally requires consistency across image sequences, motivating prior work to focus on consistency modeling.
IP-Adapter~\cite{ye2023ip} is designed to implement image prompt for pre-trained T2I diffusion models, while Cai \etal~\cite{cai2025diffusion} fine-tuned diffusion models using self-generated data through diffusion self-distillation.
ConsiStory~\cite{tewel2024training} achieved consistency by sharing internal activations, and One-Prompt-One-Story~\cite{liu2025one} enforced identity preservation by jointly encoding all prompts.
With the rise of large language models~\cite{achiam2023gpt,young2024yi}, DreamStory~\cite{he2025dreamstory} further introduced LLM-guided story generation with a consistent diffusion backbone.
Existing story generation methods achieve coherence and consistency but remain emotion-neutral. EmoStory addresses this by introducing an emotion agent and an emotion composition module.

\subsection{Affective Image Synthesis}

Affective image synthesis aims to evoke emotions through image generation or editing.
Early work mainly relied on low-level cues such as color~\cite{yang2008automatic, peng2015mixed, liu2018emotional, zhu2023emotional} and style transfer~\cite{fu2022language, sun2023msnet, weng2023affective}.
Recent studies have shown that semantic visual contents play a critical role in emotional evocation, motivating the construction of large-scale emotion datasets~\cite{yang2023emoset}.
Yang \etal~\cite{yang2023emoset} introduced EmoSet, a large-scale visual emotion dataset that laid the foundation for visual emotion research.
Building on EmoSet, EmoGen~\cite{yang2024emogen} aligns an emotion space with CLIP~\cite{radford2021learning} semantics for emotion-aware image generation.
EmotiCrafter~\cite{dang2025emoticrafter} maps discrete emotions into a continuous valence–arousal space to enable richer emotional expression.
EmoEdit~\cite{yang2025emoedit} extends emotion modeling to image editing via an emotion adapter.
EmoAgent~\cite{mao2025emoagent} further decomposes affective editing into multiple stages using collaborative agents.
While prior affective image generation methods focus on single-image emotion modeling, EmoStory extends them to coherent emotion-aware visual stories through a writer agent and a subject alignment module.
\section{Method}
\label{sec:method}

\begin{figure*}
	\centering
	\includegraphics[width=\linewidth]{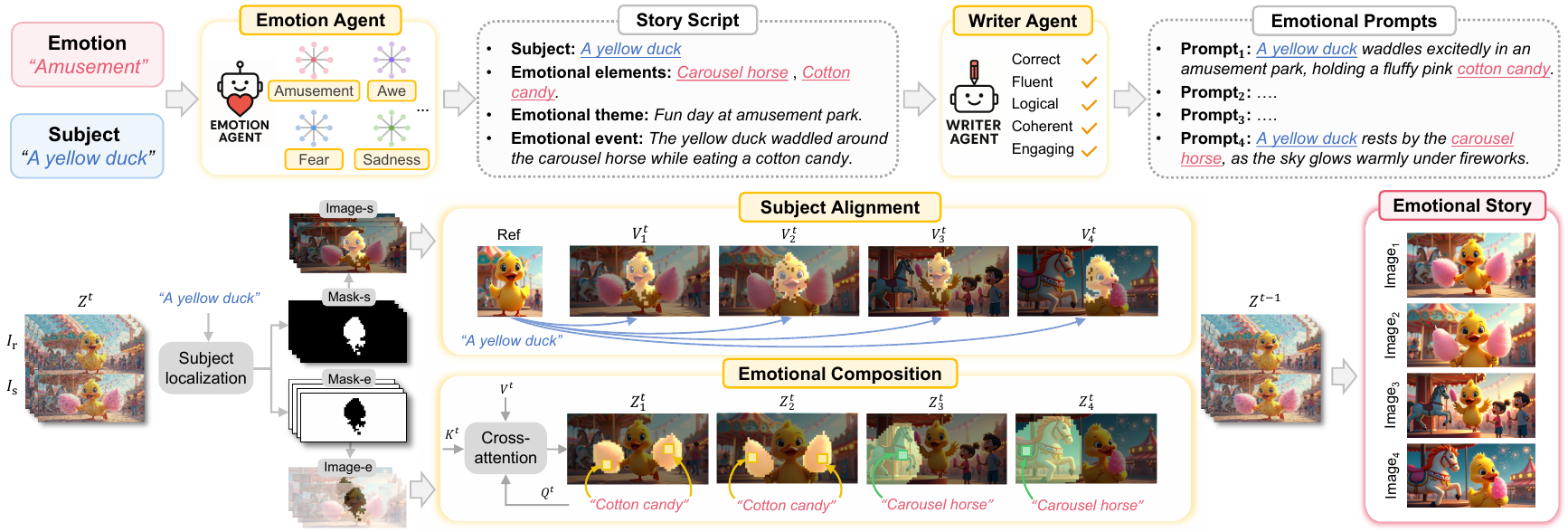}
	\vspace{-5pt}
	\caption{Overview of EmoStory. Agent-based story planning maps abstract emotions to concrete emotional prompts at semantic level, while region-aware story generation preserves subject consistency and enhances emotional expressiveness at pixel level.
	}
	\label{fig:method}
	\vspace{-5pt}
\end{figure*}

\subsection{Agent-based Story Planning}

With the great success of large language models (LLMs)~\cite{achiam2023gpt}, they have been widely adopted for story generation~\cite{liu2024intelligent, dinkevich2025story2board, shen2025storygpt}.
However, emotional story generation differs from traditional settings, as it requires grounding abstract emotion words into concrete descriptions.
To address this, we leverage the reasoning and planning capabilities of agent-based methods~\cite{li2025mccd} and introduce an emotion agent and a writer agent to progressively construct emotion-aware narratives.

\paragraph{Emotion Agent}

To build emotion agent, emotion-specific knowledge is required.
Thus, we construct eight emotion factor trees from EmoSet~\cite{yang2023emoset}, a large-scale visual emotion dataset.
In each tree, the root denotes an emotion word (\textit{Amusement}), and the leaves denote visual elements (\textit{Christmas tree}) associated with that emotion.
To ensure reliability, we remove elements that are non-emotional or rarely observed.
More details are provided in the supplementary materials.
Given an emotion word, we introduce the emotion agent to identify the corresponding emotion factor tree and select the visual elements most coherent with the provided subject.
The story script $S$ is consequently generated with the given subject, the selected emotional elements, and the summarized emotional theme and event:
\begin{equation}
	S = \{ S_{sub}, S_{ele}, S_{the}, S_{eve} \} = \mathrm{Agent}_{emotion}(c, e, \mathcal{T}_e),
\end{equation}
where $S = \{ S_{sub}, S_{ele}, S_{the}, S_{eve} \}$ denotes the generated story script, $c$ is the given subject, $e$ is the target emotion, and $\mathcal{T}_e$ represents the emotion factor tree associated with $e$.

\paragraph{Writer Agent}

To generate fluent, coherent and engaging narratives from the story script, we further introduce another agent, \ie, the writer agent, defined as:
\begin{equation}
	P = \{ p_1, p_2, p_3, p_4 \} = \mathrm{Agent}_{writer}(S),
\end{equation}
where $P = \{ p_1, p_2, p_3, p_4 \}$ represents the emotional prompts.
As shown in Fig.~\ref{fig:method}, all prompts share the same subject (\textit{A yellow duck}) for consistency, while their varied emotional elements (\textit{Carousel horse, Cotton candy}) enrich expressiveness, together yielding coherent narrative prompts.

\subsection{Region-aware Story Generation}

Emotional prompts often include rich visual elements that may spatially overlap with the subject, causing naive generation to distort the subject or weaken emotional cues.
For instance, a yellow duck and a carousel horse may unintentionally merge into a ``carousel duck'', leading to visual confusion.
To avoid such interference, we design a region-aware method that controls the generation process at the pixel level, preserving both subject consistency and emotional clarity.

\paragraph{Region Disentanglement}

Following prior work~\cite{dinkevich2025story2board, shin2025large}, we partition the image into a reference image $I_r$ and a story image $I_s$ to shared subject information in Fig.~\ref{fig:method}.
However, existing methods typically rely on cross-image attention, overlooking the fact that textual prompts can guide the generation process.
As illustrated in Fig.~\ref{fig:img-mask}, cross-image attention alone often results in subject localization drift  or causing structural blur.
Moreover, when subjects and elements are spatially close, subjects dominate the attention distribution, suppressing element expression.
To address these issues, we introduce a region-aware method that disentangles the pixel spaces of the subject and emotional elements.

Following MM-DiT~\cite{flux2024, esser2024scaling}, the story prompts, reference image, and story image are concatenated to form the diffusion input.
For simplicity, we use $Z$ to denote $Z_t$ in the following equations.
The cross-attention is subsequently computed as:
\begin{equation}
	{Q}=ZW_q, \ {K} = ZW_k, \ V = ZW_v,
\end{equation}
where $W_q$, $W_k$ and $W_v$ are learnable parameters and $Z = [P, I_r, I_s]$.
To locate subject region, we compute cross-attention between subject prompt $S_{sub}$ and story image $I_s$:

\begin{equation}
	{{A}_{sub}} = \operatorname{Softmax}\!\left(
	\frac{Q_{sub} K_{s}^\top}{\sqrt{d}}
	\right),
\end{equation}
where $Q_{sub}$ and $K_{s}$ are encoded from $S_{{sub}}$ and $I_s$. 

As a binary mask is essential to disentangle subject and element regions, we convert the attention map $A_{{sub}}$ into:
\begin{equation}
	\begin{aligned}
	M_{sub} &=
	\begin{cases}
		1, & A_{sub} \ge \tau \\
		0, & \text{otherwise}
	\end{cases}, \\
	M_{ele} &= 1 - M_{sub}, 
	\end{aligned}
\end{equation}
where $M_{sub}$ and $M_{ele}$ denotes the mask for subject region and element region respectively, $\tau$ denotes the binary  threshold.

\paragraph{Subject Alignment}

We use the subject mask to enhance an accurate subject alignment.
To preserve subject consistency, soft value mixing strategy~\cite{dinkevich2025story2board} allows smooth feature blending between reference and story images without rigid spatial boundaries.
However, relying solely on image–image cross-attention is suboptimal, so we incorporate $M_{sub}$ to inject the semantic information encoded in the textual prompts:
\begin{equation}
	V_{s}' = \lambda V_{s}\odot M_{sub} + (1-\lambda)V_{r} \odot M_{ref},
\end{equation}
where $V_s$ and $V_r$ are encoded features of $I_s$ and $I_r$, $\lambda$ is a hyper-parameter. Notably, $M_{{sub}}$ and $M_{{ref}} = RA(I_r, M_{{sub}})$ denote the subject location in the story image and reference image, where $RA(\cdot)$ is the mutually attended regions~\cite{dinkevich2025story2board}.

\begin{figure*}
	\centering
	\includegraphics[width=\linewidth]{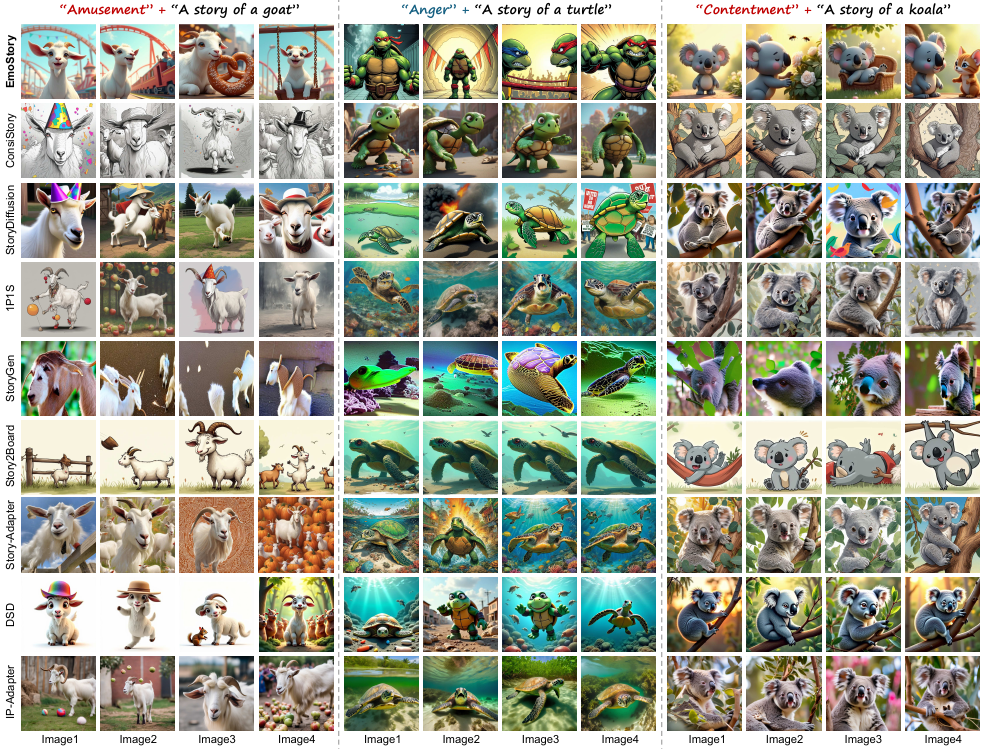}
	\vspace{-5pt}
	\caption{Comparisons with state-of-the-art methods. EmoStory is superior in both emotion evocation and story expressiveness.
	}
	\label{fig:img-sota}
	\vspace{-5pt}
\end{figure*}

\paragraph{Emotional Composition}

To place emotional elements only in non-subject region, we use the element mask $M_{ele}$ to guide the generation process.
Following prior work~\cite{lv2025rethinking, zhang2024enhancing}, we introduce an enhancement factor $\alpha$, to encourage emotional elements to be constrained to non-subject region, which value is further ablated in the supplementary materials:

\begin{equation}
	A_{{ele}} = \operatorname{Softmax}\!\left(
	\frac{Q_{{ele}} K_{s}^\top}{\sqrt{d}}
	+ \alpha M_{ele} \right),
\end{equation}
where $Q_{ele}$ and $K_{s}$ are encoded from $S_{{ele}}$ and $I_s$.
Emotional elements are encouraged to focus on non-subject regions, ensuring both emotional expressiveness and structural integrity.

\section{Experiments}
\label{sec:exp}

\subsection{Dataset and Evaluation}
\label{sec:evaluation}
\paragraph{Dataset}
EmoStory is an initial attempt at emotional story generation and adopts a training-free design.
We therefore construct an evaluation set of 25 distinct subjects generated by ChatGPT, covering both humans (\textit{A young man}) and animals (\textit{A white rabbit}).
For each subject, we define 8 emotion directions and generate 3 stories per emotion–subject pair, resulting in 600 stories, each consisting of 4 images.
This setup ensures a fair comparison across methods.

\paragraph{Evaluation Metrics}
Emotion-aware story generation is inherently a dual-target task: evoking viewers’ emotions while generating expressive visual stories.
Accordingly, we evaluate EmoStory using three metrics: Emotion Accuracy, Subject Consistency, and Prompt Alignment, where the first measures emotional effectiveness and the latter two assess story generation quality.
Emotion Accuracy measures how well the generated stories match the target emotion with a pretrained emotion classifier~\cite{yang2023emoset}.
Subject Consistency measures subject coherence across story frames~\cite{fu2023dreamsim}, and Prompt Alignment measures image–prompt alignment following~\cite{lin2024evaluating}.

\begin{figure}
	\centering
	\includegraphics[width=\linewidth]{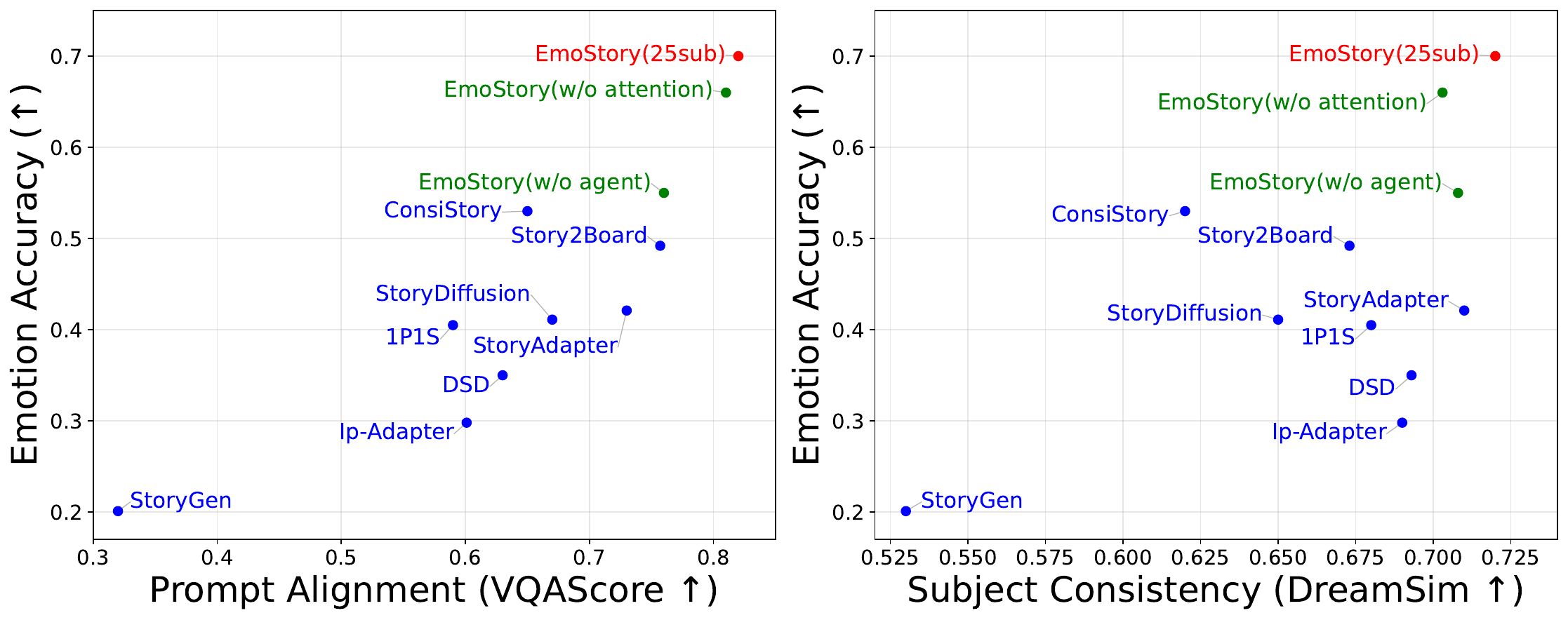}
	\vspace{-5pt}
	\caption{EmoStory (\textit{red}) outperforms all compared methods (\textit{blue}) and ablations (\textit{green}) across three evaluation metrics.
	}
	\label{fig:exp-sota}
	\vspace{-5pt}
\end{figure}

\subsection{Comparison} 
As the first work on emotion-aware story generation, we compare EmoStory with eight state-of-the-art story generation methods.
These include consistent text-to-image methods: ConsiStory \cite{tewel2024training}, StoryDiffusion \cite{zhou2024storydiffusion}, One-Prompt-One-Story (1P1S) \cite{liu2025one}, Story-Adapter \cite{mao2024story} and Story2Board \cite{dinkevich2025story2board}, the autoregressive method: StoryGen \cite{liu2024intelligent}, and the personalized generation methods: Diffusion Self-Distillation (DSD) \cite{cai2025diffusion} and IP-Adapter \cite{ye2023ip}, ensuring a comprehensive evaluation.

\paragraph{Qualitative Comparison}
We present qualitative results in Fig.~\ref{fig:img-sota}.
EmoStory outperforms other methods in both emotion evocation and story expressiveness.
Most compared methods lack explicit emotional knowledge, resulting in largely neutral visual elements and weak emotional interactions between the subject and its surroundings.
ConsiStory, StoryDiffusion, 1P1S, Story-Adapter, and Story2Board show limited emotional understanding and mainly rely on local cues, often failing to produce coherent multi-frame stories.
StoryGen struggles with complex emotional scenes and frequently produces artifacts.
Personalized methods such as DSD and IP-Adapter over-enforce identity preservation, leading to background collapse or copy-and-paste effects that hinder emotional expression.
In contrast, EmoStory generates emotionally expressive stories by introducing emotion-aware elements and interactions, owing to the effective design of emotion agent and emotional composition module.
Taking the \textit{koala} as an example, for a \textit{contentment} story, we introduce sunshine, flowers, a sofa, and friends as emotional elements while preserving subject consistency across frames in Fig.~\ref{fig:img-sota}.

\paragraph{Quantitative Comparison}
As shown in Fig.~\ref{fig:exp-sota}, EmoStory consistently outperforms all compared methods across three evaluation metrics.
The left plot illustrates the trade-off between Emotion Accuracy and Prompt Alignment, while the right plot shows Emotion Accuracy versus Subject Consistency.
There exist a naturally trade-off between emotional expressiveness and story quality: emphasizing one often weakens the other.
EmoStory effectively balances the two in both settings.
In contrast, most compared methods struggle to achieve this balance, particularly in emotion accuracy.
StoryGen fails to handle complex story prompts and express emotions, often producing collapsed images, as shown in both Fig.~\ref{fig:exp-sota} and Fig.~\ref{fig:img-sota}.
DSD and IP-Adapter over-enforce identity preservation, neglecting emotional elements and scene composition, which leads to weak emotion evocation.
ConsiStory, StoryDiffusion, 1P1S, Story-Adapter, and Story2Board capture emotions to an extent, but their emotional expression is weak and biased, as shown in Fig.~\ref{fig:img-sota}.
Overall, EmoStory achieves the best balance between emotional expressiveness and story quality, attaining 70.17\% Emotion Accuracy, 82.06\% Prompt Alignment, and 71.70\% Subject Consistency.

\paragraph{User Study}
We conduct a user study to evaluate the human-perceived performance of EmoStory.
We invite 38 participants from diverse age groups, with each session lasting at least 15 minutes.
The study includes 24 story sets, each comparing four methods: StoryGen, 1P1S, DSD, and EmoStory.
Participants are asked to evaluate each story set on subject consistency and emotion evocation.
As shown in Table~\ref{tab:table2}, EmoStory achieves the best performance across all aspects.
Specifically, EmoStory attains the highest scores in emotion evocation (74.23\%) and subject consistency (68.42\%), and also achieves the best balanced score across both aspects (83.8\%), validating its strong human alignment.

\begin{table}[t]
    \centering
    \caption{
        User preference study. The numbers indicate the percentage of participants who vote for the result.
    }
    \scriptsize
    \resizebox{\columnwidth}{!}{%
        \begin{tabular}{lccc}
            \toprule
            Method & Emotion evoking ↑ & Subject consistency ↑ & Balance ↑ \\
            \midrule
            StoryGen \cite{liu2024intelligent} & 3.29 $\pm$ 7.16\% & 3.40 $\pm$ 6.40\% & 1.39 $\pm$ 6.11\% \\
            1P1S \cite{liu2025one} & 8.22 $\pm$ 7.37 \% & 10.31 $\pm$ 11.89\% & 3.40 $\pm$ 6.01\% \\
            DSD \cite{cai2025diffusion} & 14.25 $\pm$ 10.67\% & 17.87 $\pm$ 14.77\% & 11.42 $\pm$ 11.06\% \\
            \textbf{Ours} & \textbf{74.23 $\pm$ 17.28\%} & \textbf{68.42 $\pm$ 22.75\%} & \textbf{83.80 $\pm$ 16.86\%} \\
            \bottomrule
        \end{tabular}
    }%
    \vspace{-0.05in}
    \label{tab:table2}
\end{table}
\begin{figure}
	\centering
	\includegraphics[width=\linewidth]{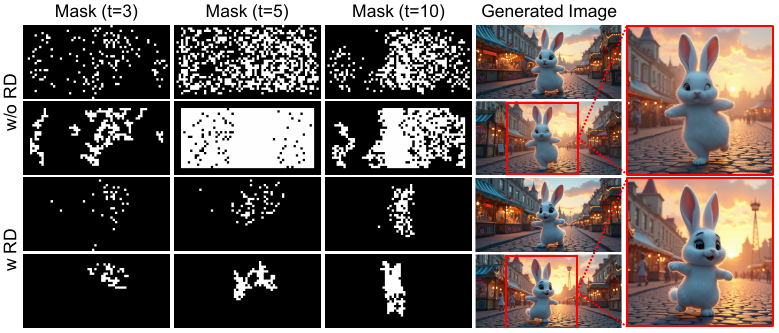}
	\vspace{-5pt}
	\caption{Visualize the evolution of subject attention masks under two settings: w/o, w region disentanglement (RD).
	}
	\label{fig:img-mask}
	\vspace{-5pt}
\end{figure}

\subsection{Visualization}
As shown in Fig.~\ref{fig:img-mask}, we visualize the evolution of subject attention masks during the generation process under two settings: without (top) and with (bottom) region disentanglement (RD).
Without RD, attention is broadly distributed over the entire image at early timesteps, causing inaccurate subject localization when conditioning only on the reference image.
As a result, value information from background regions is mixed with the subject, leading to visual artifacts such as facial collapse and limb distortion in the generated rabbit.
In contrast, with RD, EmoStory progressively suppresses non-subject regions and concentrates attention on the true subject area, yielding stable and accurate subject alignment across timesteps.
Notably, clear subject localization is achieved at t=10, which directly translates into a fine-grained and well-preserved rabbit in the final image.
These results demonstrate that region disentanglement effectively improves subject alignment and visual fidelity during story generation.

\section{Conclusion}
\label{sec:conclusion}

\subsection{Discussion}
In this paper, we present EmoStory, an emotion-aware story generation framework that produces emotionally evocative and subject-consistent narratives.
EmoStory employs an emotion agent and a writer agent in the planning stage to generate high-quality emotional prompts, and introduces region disentanglement during generation to preserve subject consistency while maintaining emotion fidelity.
Extensive experiments and user studies demonstrate the effectiveness of EmoStory across multiple evaluation metrics.

\subsection{Limitations}
EmoStory still faces several challenges.
First, EmoStory considers only eight discrete emotion categories, whereas real-world emotions are more complex.
Second, emotional elements remain limited for certain emotions (\eg, anger is frequently associated with fire), which reduces narrative diversity in generated stories.
Third, the current design focuses on a single subject input and does not yet support long prompts or multi-subject interactions.
Addressing these limitations by incorporating richer emotion representations, more diverse emotional cues, and multi-subject story modeling is an important direction for future work.

\bibliographystyle{IEEEbib}
\bibliography{EmoStory}

\vspace{12pt}
\color{red}

\end{document}